# IPS300+: a Challenging Multimodal Dataset for Intersection Perception System

Huanan Wang[a], Xinyu Zhang*[a], Jun Li[a], Zhiwei Li[a], Lei Yang[a], Yongqiang Deng[b]

[a]Tsinghua University, [b]Beijing Wanji Technology Co. Ltd

*Abstract—* **Due to the high complexity and occlusion, insufficient perception in the crowded urban intersection can be a serious safety risk for both human drivers and autonomous algorithms, whereas CVIS (Cooperative Vehicle Infrastructure System) is a proposed solution for full-participants perception in this scenario. However, the research on roadside multimodal perception is still in its infancy, and there is no open-source dataset for such scenario. Accordingly, this paper fills the gap. Through an IPS (Intersection Perception System) installed at the diagonal of the intersection, this paper proposes a high-quality multimodal dataset for the intersection perception task. The center of the experimental intersection covers an area of 3000m$^2$, and the extended distance reaches 300m, which is typical for CVIS. The first batch of open-source data includes 14198 frames, and each frame has an average of 319.84 labels, which is 9.6 times larger than the most crowded dataset (H3D dataset in 2019) by now. In order to facilitate further study, this dataset tries to keep the label documents consistent with the KITTI dataset, and a standardized benchmark is created for algorithm evaluation. Our dataset is available at: http://www.openmpd.com/column/IPS300+.**

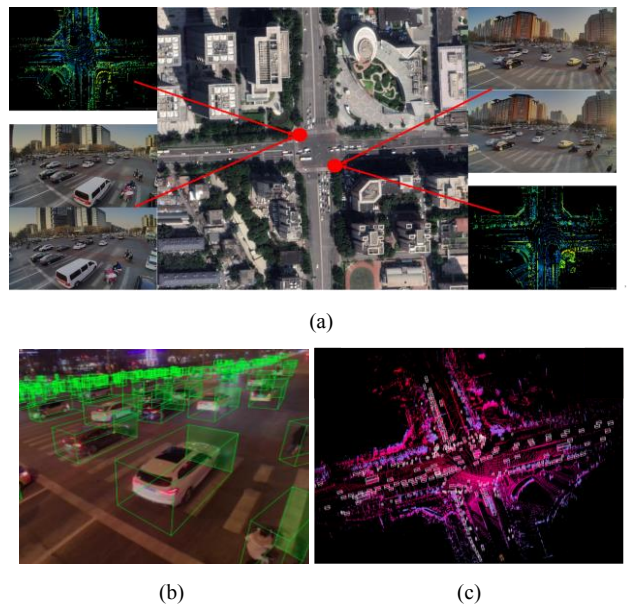

Fig. 1. A brief view of IPS300+ dataset. (a) is the aerial view of the experimental intersection. The red dots in the images represent the locations of the IPUs and following the arrows are the data collected by Lidar and binocular cameras by different IPUs. (b) is the result of 3D bounding box annotated in the point cloud and projected on the image plane. (c) is the point cloud of two IPUs after registration and the annotated 3D bounding boxes of objects.

## I. INTRODUCTION

Robust perception of the surrounding environment has always been one of the most crucial factors in autonomous driving. Lately, perception through the onboard sensing unit has been widely studied as a hot spot. A large number of datasets such as KITTI [1], nuScenes [2] and Waymo [3] have greatly promoted related researches. However, when dealing with extremely complex and severely obstructed scenes such as crowded urban intersections, even experienced human drivers cannot ensure the safety of driving simply based on the information seen inside the car, let alone autonomous algorisms. ERSO2018 (European Road Safety Observatory) public traffic accident statistics report [4] shows that, in 2016, of the 9693 urban traffic accidents that occurred in Europe, 3839 occurred at intersections, accounting for 39.6% of the total urban accidents.

Therefore, driving safely across urban intersections remains a challenging open question for autonomous driving. Recently, CVIS has attracted broad attention in both academia and industry. As a solution to insufficient perception in large-scale urban scenes, CVIS bridges the gap between smart transportation and smart vehicles, and could become an important infrastructure for the future smart city. In CVIS, RSUs (Road-Side Units) are installed to get reliable perception of the entire intersection from the top-down perspective, and the perception results are sent to the passing vehicles through V2I.

Among all the pilot studies in various countries, Ko-FAS project [5-7] sponsored by EU established an IPS in an intersection of Aschaffenburg. RSUs and OBUs enable vehicles to get the whole view of the intersection and the ADAS (Advanced Driver Assistance Systems) can warn the drivers of potential collision at blind spots. CICAS-SSA project [8] sponsored by US DOT focused on the warning for incoming vehicles in rural roads and told drivers the correct time to cut in. To achieve this, 16 Radars and 8 Lidars were installed as RSUs and the whole IPS costs 180890 dollars in 2010. However, it is difficult to promote the approach due to the unaffordable price. In J-Safety project [9], the UTMS of Japan used the perception information of several intersections to adjust the dynamics of vehicles to promote the fuel economy. However, these projects do not take the advantage of deep learning technology in roadside perception tasks and there is no open-source multimodal dataset for large-scale urban intersection yet. The data-driven character of deep learning method makes the data even more important for multimodal roadside perception research.

In this paper, we proposed the largest objects-per-frame multimodal dataset IPS300+ for roadside perception in urban intersections, aiming to promote the research for 3D target detection by roadside units in CVIS. The IPS300+ data is published under CC BY-NC-SA 4.0 license.

* Corresponding author. Email addresses: xyzhang@tsinghua.edu.cn

The main contributions of this paper are as follows:

- The first multimodal dataset (including point clouds and images) available for roadside perception tasks in large-scale urban intersection scene. The point clouds remain usable within 300m.

- The most challenging dataset with the highest label density. The proposed dataset includes 14198 frames of data and every single frame has an average of 319.84 labels, which is 65.3 times larger than KITTI [1].

- The 3D bounding box is labeled at 5Hz, which provides dense truth data for 3D target detection task and the coming tracking task.

- A feasible and affordable solution for IPS construction and a wireless approach for time synchronization and spatial calibration are provided. The open questions for algorisms are also mentioned in this paper.

## II. RELATED WORKS

Since there is hardly any dataset that focuses on roadside perception in urban intersections, the related works are separated into three sections which cover all related datasets from autonomous driving to traffic monitoring.

### A. On-board Perception Datasets for Autonomous Driving

The datasets for on-board perception task are relatively abundant when comparing with roadside perception, and datasets with both point clouds and images are enriched in recent years. The KITTI dataset released by A. Geiger et al. [1] has been widely used as the benchmark for algorism evaluation. Besides, H. Caesar et al. [2] released the nuScenes dataset containing 40k labeled data; Google Waymo released their dataset [3] containing 12M labeled data; OEMs (Original Equipment Manufacturers) released their dataset either such as Audi's A2D2 dataset [10] and Honda's H3D dataset [11].

Among all these datasets, Lidars and cameras are installed on different locations of the car with the labeling frequency ranging from 2Hz to 10Hz. All these precious labeled data have greatly promoted the research of onboard perception for autonomous driving, and the algorisms driven by these data have made a remarkable result.

### B. Roadside Traffic Monitoring Datasets

Most of the existing datasets collected by roadside units are based on images and 2D bounding boxes are provided for the targets. Among them, NGSIM [12, 13] released in 2007 is one of the most widely used datasets in this research area. This dataset contains the traffic data of US highway 101 and interstate 80 freeway collected by FHWA DOT. The sensing cameras were installed on the top of the building tens of meters above the ground.

Besides NGSIM, Aachen University of Technology released the HighD dataset [14] in 2018 which covered the traffic scene of German highway. The data were collected by a camera mounted on a drone. BUPT released their roadside re-identification dataset VeRi [15] in the same year, which was captured by 20 road surveillance cameras. However, they only provide the cut images of the targets separately instead of the whole image of the camera.

### C. Roadside Target Perception Datasets

The only reachable multimodal dataset for the intersection scene is from the Ko-FAS project [5-7] in 2014. It contains 6min 28s (4850 frames) data from an intersection of a town in Aschaffenburg. The widths of the branches are 20m and 15m. Sensors include 14 8-layer Lidars and 8 cameras (only 2 available) are installed 5-7m above the ground to get the full view of the road. The sensors and data details are shown in Table I. Fig. 2 shows a single frame of this dataset. Due to limited points of Lidars, the bus 50m from the center of the intersection only has 11 points, which is hardly recognizable for both humans and algorisms.

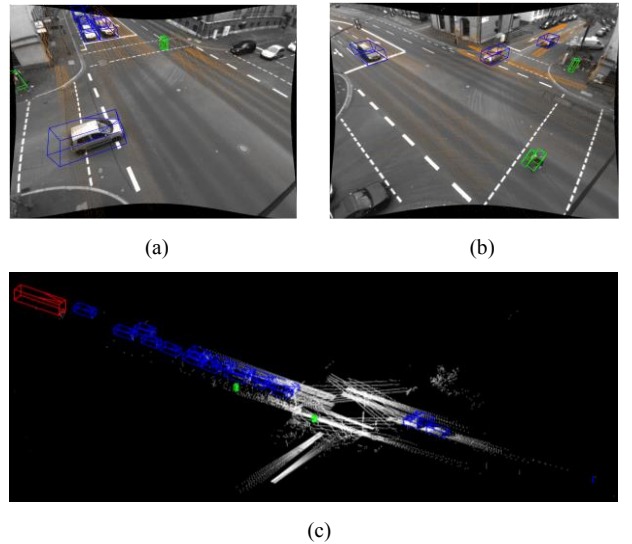

Fig. 2. A single frame of Ko-FAS dataset. (a) and (b) are two viewpoints of the available cameras. The orange points in the images are the point cloud projected on the image planes. (c) is a point cloud after registration of 14 well-synchronized Lidars. The red bounding box represents a bus 50m from the center of the intersection which only has 11 Lidar points left.

## III. IPS300+ DATASET

### A. Data Collection Platform

While cameras, Lidars, and Radars are widely used for onboard perception tasks, it is still an open question what kind of roadside sensors can meet the requirements of CVIS. There has no typical dataset and unified evaluation standard for this task, neither. This paper provides a possible solution for IPS by learning from the onboard sensors and making tentative adjustments for roadside perception tasks.

This paper takes IPU (Intersection Perception Unit) as the smallest unit of the task. As shown in Fig. 3, each IPU includes one 80-layer Lidar with specially designed vFOV, two 5.44MP color cameras, and one GPS. The Lidar is installed horizontally to the ground, while the cameras are tilted downward at 30° angle. The detailed parameters of related sensors are as follows:

- A Robosense Ruby-Lite Lidar: 80 beams; 10Hz; detect range: 230m; single echo mode; hFOV: 0°-360°; vFOV: -25°-15°; angle resolution: 0.2°; 144000 points in a single frame. (The Lidar beam is unevenly distributed in the horizontal. The dense part is in the middle which scans the opposite branches of the intersection.)

- Two Sensing-SG5 color cameras. 30Hz; 5.44MP; CMOS: Sony IMX490RGGB; rolling shutter; HDR dynamic range: 120dB; image size: 1920×1080. The two cameras form a binocular system with a distance of 1.5m.

- A BS-280 GPS used for positioning and time sync, timing error: less than 1 μs. (Since IPU are static in WGS84 coordinate system, the GPRS output is set to 1Hz.)

- An industrial computer for data collection. CPU: Intel I7-10700; Memory: 16G; Ubuntu 20.04; ROS Noetic.

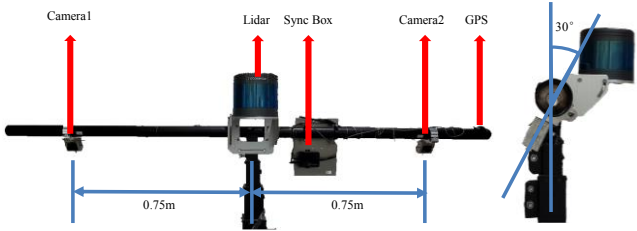

Fig. 3. Sensors location on each IPU

### B. Location and Sensors Setup

The intersection of Chengfu Road and Zhongguancun East Road is chosen for data collection in Haidian District, Beijing. The central area of the intersection reaches 60 m×50m and suffers heavy occlusion problems because of the large traffic volume. Since there are lots of universities nearby, the number of pedestrians, cyclists and tricycles is relatively larger than those of other locations. All these factors make this intersection a challenging driving scene, which is also a typical scenario for CVIS.

After weighing all the factors including maximizing sensing range, minimizing traffic impact, economy friendly and compatibility with existing facilities, two IPUs were equipped at the diagonal of the intersection 5.5m from the ground. The installation locations of two IPUs are shown in Fig. 1.

### C. Sensors Calibration

**Time synchronization:** this part includes the synchronization of each sensor inside one IPU and synchronization between different IPUs. GPS-PPS signals are used to get the unified time.

TABLE I. THE DATA DETAILS OF IPS300+ AND KO-FAS DATASET FOR COMPARE

|  | IPS300+ | Ko-FAS |
|---|---|---|
| **Intersection Size** | 60m×50m | 20m×15m |
| **Frame** | 14198 | 4850 |
| **Anno. Frequency** | 5Hz | 12.5Hz |
| **Sensors Details** | | |
| *Camera* | 4×Sensing-SG5 | 8×Baumer TXG-04 (2 available) |
| *Lidar* | 2×Robosense Ruby-Lite | 14×SICK LD-MRS 8 |
| *GPS* | 1×BS-280 | --* |
| **Data Details** | | |
| *Images* | 4×1920×1080 | 8×656×494 |
| *Point clouds* | 2×144000 | 14×1200 |

*No GPS are used in Ko-FAS for sensors calibration

Each sensor on the same IPU is connected to the GPS-PPS output port of GPS through a wire. After the PPS trigger sends out, the Lidar rotates to 0°, which is perpendicular to the crossbar of IPU, the camera triggers its shutter. Due to the constraint of the crossbar, it is guaranteed that when the Lidar

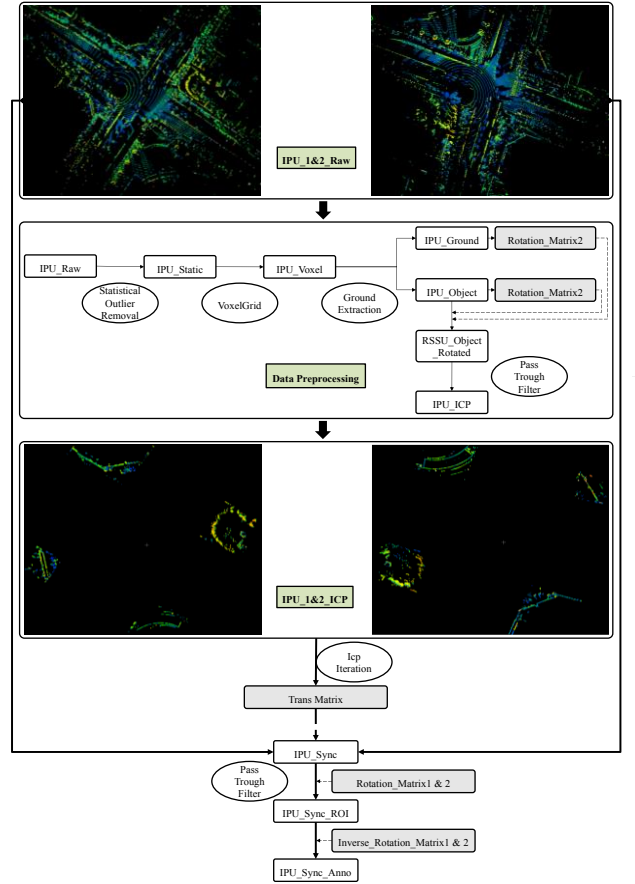

Fig. 4. Pipeline of the spatial calibration between IPUs. (The details for parameters setting can be downloaded from the dataset website.)

rotates to the center view of the cameras, the cameras open their shutters and take one photo.

The key to maintaining time synchronization between two IPUs is to unitize the time trigger. Since the upper limit of the timing error of GPS is 1μs, the cumulative error between two IPUs is less than 2μs, which is acceptable when compared with the frequency of sensors.

**Spatial calibration:** similarly, the spatial calibration includes the same two parts.

Within the IPU, the method proposed in [16] is adopted to get the distortion and internal matrix of each camera. The calibration of binocular cameras uses the method proposed in [17]. The external parameters between camera and Lidar are calculated by using the method proposed in [18, 19]. Since these processes are basically the same with the calibration process used in autonomous vehicles, we do not elaborate on them.

Between IPUs, the spatial calibration problem can be treated as a registration problem of the point clouds from different IPUs. However, the distance between the two sensing devices has reached 70m, and the point cloud features are relatively sparse. ICP-based methods and NDT cannot converge if the raw point clouds are directly used. In order to minimize the registration error, manually screening of the usable feature is needed before registration. Fig 4 shows the pipeline of the point clouds registration process in this paper.

Since two Lidars are far apart, the point clouds features representing the same location may be different (as shown in Fig. 5). In order to ensure the consistency of the features from the point clouds, only the buildings are selected for ICP iterations. Specifically, after conventional operations such as statistical outlier filter and voxel grid filter, the method proposed in [20] is used to initially divide the point cloud into ground and targets. Then the ground points are fitted to a plane by RANSAC [21] and rotated into the XY plane of the intersection coordination. The transformation matrix is recorded as Rotation_Matrix1. After Rotation_Matrix1, the height of the object is only coupled to z axis. Then, the barrier in the middle of the road is manually selected and rotated to x axis by Rotation_Matrix2. The branches of two roads are coupled to x and y axes separately. Then, the pass-through filter is used to select the building and sent them to ICP algorithm. And the output of ICP is used as the Trans_matrix from IPU1 to IPU2.

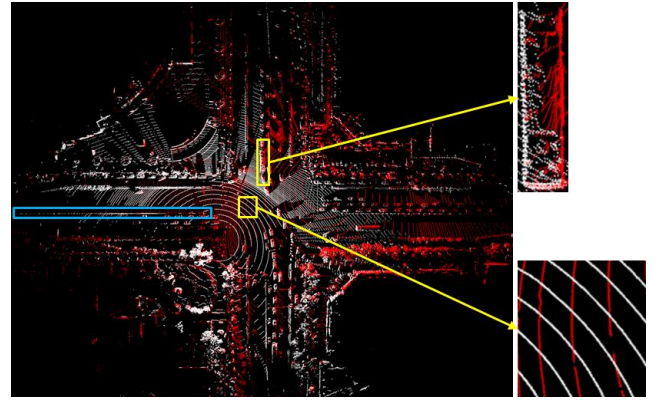

Fig. 5. A single-frame of point cloud after registration. In which the red points and the white points are from IPU1 and IPU2 respectively. The details on the right side indicate the difference of point clouds at the same position due to the distance of two IPUs. The blue box is the manually selected feature for solving Rotation_Matrix2.

### D. Data Specification

IPS300+ includes data that covers different times of the day in 14198 frames of data. The point clouds are stored as a single registered .pcd file (PCD_Sync_Anno) in each frame for labeling, and the size of a single frame is about 8M (4M×2). The images are stored in .png format and are about 12M (3M×4) per frame. Since there are hundreds of stationary vehicles parked on the sidewalks at the intersection and have no effect on the CVIS research, we cut the registered point cloud by Rotation_Matrix and pass-through filter. Only the targets on the lanes are remained for the labeling process (as shown in Fig. 4). The detailed descriptions of our dataset are shown in Table I.

The raw data are available at IPU1 and IPU2 in data folders. Each IPU folder including the point clouds data and binocular camera data, related parameters of sensors can be found in calib_file.txt. In order to flexible the data processing, this paper also provides the registered filtered point clouds in PCD_Sync_Anno folder (the point clouds coordinate is consistent with IPU1 point clouds and the 3D labels are in the same coordinate). This data structure provides the greatest convenience possible for the IPS algorithm research.

It is worth mentioning that besides the problem of multimodal fusion in the same IPU, the multi-spatial fusion of

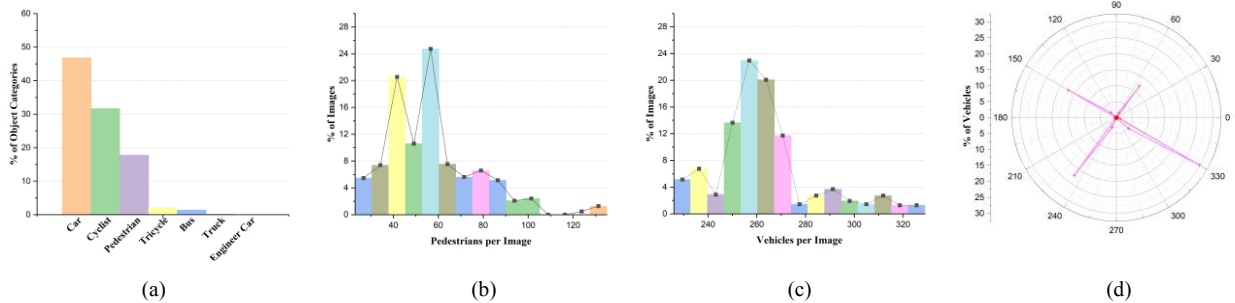

Fig. 6. The Statistics of IPS300+ dataset. (a) is the distribution of different categories. (b) and (c) respectively show the statistical result of the number of pedestrians and vehicles in different frames. (d) is the orientation of the vehicles.

TABLE II. THE DATA DETAILS OF IPS300+ AND KO-FAS DATASET FOR COMPARISON

| Count | Ko-FAS | KITTI | A2D2 | nuScenes | Waymo OD | Honda H3D | our IPS300+ |
|---|---|---|---|---|---|---|---|
| Average per frame | | | | | | | |
| *pedestrians* | 3.7 | 0.8 | 0.35 | 7.0 | 12.2 | 16.5 | **56.8** |
| *vehicles* | 16.0 | 4.1 | 3.0 | 20.0 | 26.5 | 16.9 | **263.0** |

Since the datasets differ in the classification of vehicles, the column of 'vehicles' in this table includes but not limited to cars, cyclists, tricycles and buses.

different IPUs is also a novel but significative problem for CVIS. All these problems can be studied under our IPS300+ dataset.

### E. Ground Truth Labels (Statistics)

The first batch of open source labeled data includes 623 frames in the evening rush hour, which has the most intensive traffic flow and the highest risk of accidents. The remaining data are published as unlabeled data, and the labels will be opened for the public upon the release of this paper. The labeled data include 7 categories: pedestrian, cyclist, tricycle, car, bus, truck and engineering car. In order to facilitate the data processing progress, this paper provides a label document consistent with KITTI, and some adjustments have been made according to the special scenario of our dataset. The details can be found on our website.

The labeling task uses Lidar point clouds and related images at the same time. The 3D bounding box of the target is selected in the point clouds with the help of reference images, and the bounding box is projected to each camera plane through the external parameter. Fig. 6 shows some statistical results of labeled targets. 6 (a) shows that in the intersection of IPS300+ dataset, the main traffic participants are cars, cyclists and pedestrians. 6 (b) and 6 (c) show that the number of pedestrians and vehicles are about 40-60 and 250-270 per frame separately. The on-board perception system is impossible to guide autonomous vehicles under such numbers of targets due to the limited perspective and computing power, which makes the CVIS indispensable. 6(d) shows that the orientations of vehicles are mostly in the direction parallel to the two branches of the intersection, which can be an important priori knowledge for the design of intersection perception algorisms.

Table II shows the comparison result of IPS300+ and other open-source datasets. The average pedestrians and vehicles in each frame of our dataset reach 56.8 and 263.0, which exceeded over 3.4 times and 9.9 times than the largest existing dataset respectively. The severe occlusion problem in onboard perception makes this intersection a typical scenario for CVIS study.

## IV. EXPERIENCE

Target detection is one of the most challenging tasks in the research of autonomous perception. Two main tasks are provided in this paper as Lidar-based detection and Camera-based detection (tracking tasks will also be available after ID checking). For the 3D target detection task, this paper uses the AP metric (including $AP_{3D}$, $AP_{BEV}$) consistent with KITTI [1, 22] to evaluate the performance of the detection algorithm. The IoU threshold is set to 0.25 for pedestrians, 0.5 for cyclists, and 0.7 for the other five types of vehicles.

### A. 3D Detection by Lidar

After sufficient research of the existing method of Lidar-based target detection, this paper uses PointPillars [23] as a baseline because of its high efficiency and accuracy. In order to quantitatively analyze the coverage distance and the accuracy of the IPS (both hardware and algorisms as a perception system) for the actual intersection scenario, this paper uses 50m as an interval to count the targets detection results in different distances from the center of the intersection, i.e. the midpoint of two IPUs. A map for the accuracy of vehicle detection within distances and a typical failure case are shown in Fig. 7. The quantitative results are summarized in Table III.

It can be seen that for the vehicle detection task, the accuracy of target detection decreases by distance. The $AP_{3D}$ at a distance of 150 meters stays at 66.81%, i.e. the range of vehicle perception reaches about 15 vehicles after the stop lines of the intersection. This top-down view of the intersection will provide rich information for the related function design such as intersection collision warning, vehicle

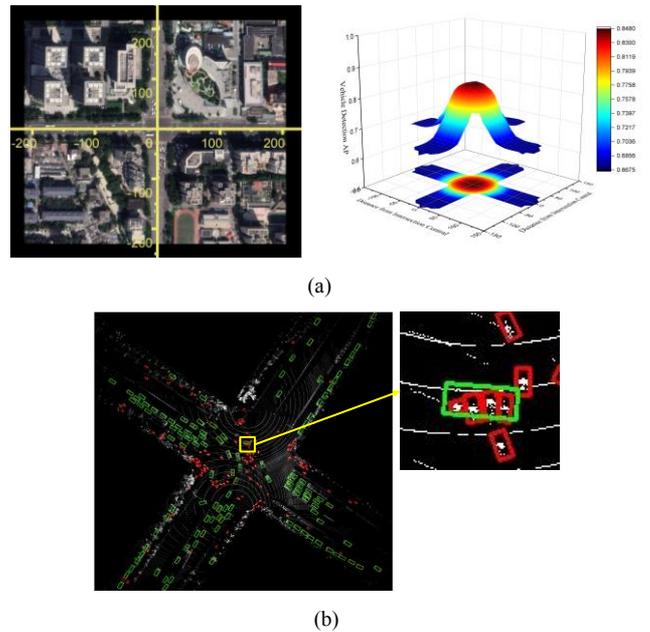

(a)

(b)

Fig. 7. The 3D Detection result by PointPillars. (a) shows the AP value of vehicle detection on the branch roads with distance. (b) is the visual result of a single frame where the green boxes represent the predict results of vehicles and the red boxes represent the groundtruth of vehicles and pedestrians. A typical error is: when pedestrians appear as a dense crowd, the detection algorism might mistakenly treat them as a vehicle.

TABLE III. 3D DETECTION BASED ON POINTPILLARS

|  |  | AP$_{3D}$ |
|---|---|---|
| **Distance from the Intersection Center** | **30m** | 82.90% |
|  | **60m** | 73.48% |
|  | **90m** | 68.05% |
|  | **120m** | 67.44% |
|  | **150m** | 66.81% |

TABLE IV. 2D DETECTION BASED ON CENTERNET

|  | AP |
|---|---|
| **Pedestrian** | 69.68% |
| **Car** | 70.89% |
| **Cyclist** | 71.10% |
| **Total** | 70.56% |

Since the label of other categories are not large enough, they are not used during the training process.

trajectory planning, decision-making in autonomous driving and traffic flow controlling, traffic light timing system in ITS.

*B. 3D Detection by Camera*

The SOTA (State of the Art) algorisms for monocular 3D target detection can be divided into two categories.

Stage-to-stage methods all assume that the XY plane of the camera coordinate system is parallel to the ground and the bottom of the bounding box is fitted on the ground, such as [24-28]. Under this assumption, the output of the network is (h, w, l, x, y, z, θ), where θ is the yaw angle, the roll angle and pitch angle are fixed to 0. However, on IPUs, the camera is tilted to get a top-down view of the whole intersection. The XY plane of the camera coordinate is not parallel to the ground. None of the roll, yaw and pitch angles of the object are 0 and these angles are changing according to the distance. Therefore, these methods require further modifications for IPU tasks.

Another SOTA way such as [29-31] generates pseudo point cloud through images, and this process requires dense depth maps for supervised training of pseudo point cloud generation net. However, the acquisition of dense deep maps is extremely difficult for IPUs since existing methods such as [32] are mainly based on the integration of adjacent frames. It works only when the Lidar is equipped on-board and moves over time, which is invalid for the Lidar stationed on IPU.

So, as far as we know, monocular 3D detection on IPUs is still an open question.

*C. 2D Detection by Camera*

This paper uses CenterNet [26] with Resnet34 as the backbone of 2D detection task and the detection result are shown in Table IV. The AP for pedestrian, car and cyclist are similar since the sample are sufficient for all categories in intersection scenarios. In actual IPS scenarios, the pedestrians on the opposite side of the intersection can be 70m or more away from the IPU, there are only about 50 pixels even if one pedestrian is not occluded. Not to mention that the pedestrians in the intersection are usually heavily occluded and act as crowds. All these problems lead to the poor performance of 2D detection by camera and there is still a large room for optimization in both hardware and algorisms.

V. CONCLUSION AND FUTURE WORK

In this paper, we present the first multimodal dataset for roadside perception in large scale urban intersection, and provide high-quality open source data for 3D detection task in CVIS. This dataset aims to provide a top-down view of the crowded intersection and accumulates the full-participation perception task through IPS. The perception results can be sent to cars that entering the intersection for beyond-visual perception and corresponding decision making & planning process.

When comparing with the existing datasets in autonomous driving, the scenario in IPS300+ is extremely complex and has severe occlusion problem, and our dataset has the highest label density so far. The current web of our data set includes 3D object detection tasks based on Lidar and cameras, and provides a ranking list of corresponding algorithms.

In the future, we will continue releasing the data of different times. After the ID checks for targets are completed, multiple ID documents in 5Hz, 20s fragments of the published data will be released for 3D multi-target tracking task.

The unlabeled data of a car entering the intersection at the same time will also be open to public upon the release of this paper. The on-board sensors include a RS-Ruby 128-layer Lidar and a Basler 2MP camera. The label documents for that data will also be published soon.

ACKNOWLEDGMENT

This work was supported by the National High Technology Research and Development Program of China under Grant No. 2018YFE0204300, and the Beijing Science and Technology Plan Project(Z191100007419008), and the Guoqiang Research Institute Project(2019GQG1010) and sponsored by Tsinghua University-Didi Joint Research Center for Future Mobility. The IPS300+ dataset was annotated by Datatang Co. Ltd.

REFERENCES

[1] A. Geiger, P. Lenz, and R. Urtasun, "Are we ready for autonomous driving? The KITTI vision benchmark suite," in IEEE Conference on Computer Vision & Pattern Recognition, 2012.
[2] H. Caesar, V. Bankiti, A. H. Lang, S. Vora, and O. Beijbom, "nuScenes: A Multimodal Dataset for Autonomous Driving," in 2020 IEEE/CVF Conference on Computer Vision and Pattern Recognition (CVPR), 2020.
[3] P. Sun et al., "Scalability in perception for autonomous driving: Waymo open dataset," in Proceedings of the IEEE/CVF Conference on Computer Vision and Pattern Recognition, 2020, pp. 2446-2454.
[4] "ERSO 2018 Annual accident report 2018," European Road Safety Observatory, Brussels, 2018.
[5] E. Strigel, D. Meissner, F. Seeliger, B. Wilking, and K. Dietmayer, "The Ko-PER intersection laserscanner and video dataset," in 17th International IEEE Conference on Intelligent Transportation Systems

(ITSC), 8-11 Oct. 2014 2014, pp. 1900-1901, doi: 10.1109/ITSC.2014.6957976.
[6] D. Meissner and K. Dietmayer, "Simulation and calibration of infrastructure based laser scanner networks at intersections," in 2010 IEEE Intelligent Vehicles Symposium, 2010: IEEE, pp. 670-675.
[7] M. Goldhammer, E. Strigel, D. Meissner, U. Brunsmann, K. Doll, and K. Dietmayer, "Cooperative multi sensor network for traffic safety applications at intersections," in 2012 15th International IEEE Conference on Intelligent Transportation Systems, 2012: IEEE, pp. 1178-1183.
[8] L. Alexander et al., "The Minnesota mobile intersection surveillance system," in 2006 IEEE Intelligent Transportation Systems Conference, 2006: IEEE, pp. 139-144.
[9] M. Sugimoto, "Driving safety support system: Dsss," in Proceedings of the IEEE International Vehicle Electronics Conference (IVEC'99)(Cat. No. 99EX257), 1999: IEEE, pp. 480-484.
[10] J. Geyer et al., "A2d2: Audi autonomous driving dataset," arXiv preprint arXiv:2004.06320, 2020.
[11] A. Patil, S. Malla, H. Gang, and Y.-T. Chen, "The h3d dataset for full-surround 3d multi-object detection and tracking in crowded urban scenes," in 2019 International Conference on Robotics and Automation (ICRA), 2019: IEEE, pp. 9552-9557.
[12] J. H. a. J. Colyar. NGSIM - Interstate 80 Freeway Dataset [Online] Available: https://www.fhwa.dot.gov/publications/research/operations/06137/06137.pdf
[13] J. C. a. J. Halkias. NGSIM - US Highway 101 Dataset. [Online] Available: https://www.fhwa.dot.gov/publications/research/operations/07030/07
[14] R. Krajewski, J. Bock, L. Kloeker, and L. Eckstein, "The highD Dataset: A Drone Dataset of Naturalistic Vehicle Trajectories on German Highways for Validation of Highly Automated Driving Systems," 2018.
[15] X. Liu, W. Liu, T. Mei, and H. Ma, "PROVID: Progressive and Multimodal Vehicle Reidentification for Large-Scale Urban Surveillance," IEEE Transactions on Multimedia, pp. 1-1, 2018.
[16] Z. Zhang, "A flexible new technique for camera calibration," IEEE Transactions on pattern analysis and machine intelligence, vol. 22, no. 11, pp. 1330-1334, 2000.
[17] D. B. Gennery, "Stereo-camera calibration," in Proceedings ARPA IUS Workshop, 1979, pp. 101-107.
[18] K. S. Arun, T. S. Huang, and S. D. Blostein, "Least-squares fitting of two 3-D point sets," IEEE Transactions on pattern analysis and machine intelligence, no. 5, pp. 698-700, 1987.
[19] L. Zhou, Z. Li, and M. Kaess, "Automatic extrinsic calibration of a camera and a 3d lidar using line and plane correspondences," in 2018 IEEE/RSJ International Conference on Intelligent Robots and Systems (IROS), 2018: IEEE, pp. 5562-5569.
[20] K. Zhang, S.-C. Chen, D. Whitman, M.-L. Shyu, J. Yan, and C. Zhang, "A progressive morphological filter for removing nonground measurements from airborne LIDAR data," IEEE transactions on geoscience and remote sensing, vol. 41, no. 4, pp. 872-882, 2003.
[21] S. Choi, T. Kim, and W. Yu, "Performance evaluation of RANSAC family," Journal of Computer Vision, vol. 24, no. 3, pp. 271-300, 1997.
[22] A. Simonelli, S. R. R. Bulò, L. Porzi, M. López-Antequera, and P. Kontschieder, "Disentangling Monocular 3D Object Detection," 2019.
[23] A. H. Lang, S. Vora, H. Caesar, L. Zhou, J. Yang, and O. Beijbom, "Pointpillars: Fast encoders for object detection from point clouds," in Proceedings of the IEEE/CVF Conference on Computer Vision and Pattern Recognition, 2019, pp. 12697-12705.
[24] Z. Liu, Z. Wu, and R. Tóth, "Smoke: single-stage monocular 3d object detection via keypoint estimation," in Proceedings of the IEEE/CVF Conference on Computer Vision and Pattern Recognition Workshops, 2020, pp. 996-997.
[25] Y. Tang, S. Dorn, and C. Savani, "Center3D: Center-based Monocular 3D Object Detection with Joint Depth Understanding," arXiv preprint arXiv:2005.13423, 2020.
[26] X. Zhou, D. Wang, and P. Krähenbühl, "Objects as points," arXiv preprint arXiv:1904.07850, 2019.
[27] P. Li, "Monocular 3D Detection with Geometric Constraints Embedding and Semi-supervised Training," arXiv preprint arXiv:2009.00764, 2020.
[28] P. Li, H. Zhao, P. Liu, and F. Cao, "RTM3D: Real-time monocular 3D detection from object keypoints for autonomous driving," arXiv preprint arXiv:2001.03343, vol. 2, 2020.
[29] Y. Wang, W.-L. Chao, D. Garg, B. Hariharan, M. Campbell, and K. Q. Weinberger, "Pseudo-lidar from visual depth estimation: Bridging the gap in 3d object detection for autonomous driving," in Proceedings of the IEEE/CVF Conference on Computer Vision and Pattern Recognition, 2019, pp. 8445-8453.
[30] X. Weng and K. Kitani, "Monocular 3d object detection with pseudo-lidar point cloud," in Proceedings of the IEEE/CVF International Conference on Computer Vision Workshops, 2019, pp. 0-0.
[31] R. Qian et al., "End-to-end pseudo-lidar for image-based 3d object detection," in Proceedings of the IEEE/CVF Conference on Computer Vision and Pattern Recognition, 2020, pp. 5881-5890.
[32] J. Uhrig, N. Schneider, L. Schneider, U. Franke, T. Brox, and A. Geiger, "Sparsity invariant cnns," in 2017 international conference on 3D Vision (3DV), 2017: IEEE, pp. 11-20.